\documentclass[]{template}

\usepackage[utf8]{inputenc}             
\usepackage[T1]{fontenc}                
\usepackage{url}                        
\usepackage{booktabs}                   
\usepackage{multirow}
\usepackage{colortbl}
\usepackage{multicol}
\usepackage{amsfonts}                   
\usepackage{nicefrac}                   
\usepackage{microtype}                  
\usepackage[dvipsnames]{xcolor}         
\usepackage{algorithmic}
\usepackage{latexsym}
\usepackage{natbib}

\usepackage{graphicx}
\usepackage{float}
\usepackage{subcaption}
\usepackage{wrapfig}
\usepackage{lipsum}

\usepackage{bm}

\usepackage{tabularx} 
\usepackage{ragged2e} 
\newcolumntype{L}{>{\RaggedRight\hangafter=1\hangindent=0em}X}

\usepackage{enumitem}

\usepackage{amsmath}
\usepackage{amssymb}
\usepackage{mathtools}
\usepackage{amsthm}
\usepackage{authblk}

\setboolean{logo}{true}    

\usepackage[linesnumbered,ruled,vlined]{algorithm2e}

\hypersetup{
    colorlinks=true,
    linkcolor=red,
    citecolor=Cerulean,
    filecolor=magenta,      
    urlcolor=magenta,
}

\usepackage[capitalize,noabbrev]{cleveref}
\crefname{section}{§}{§§}
\Crefname{section}{§}{§§}

\usepackage{calligra}
\DeclareMathAlphabet{\mathcalligra}{T1}{calligra}{m}{n}

\usepackage{pifont}

\theoremstyle{plain}

\theoremstyle{definition}

\theoremstyle{remark}

\renewcommand{\paragraph}[1]{\vspace{1mm}\noindent\textbf{#1}}

\DeclareCaptionLabelFormat{cont}{#1~#2\alph{ContinuedFloat}}
\usepackage[most]{tcolorbox}
\tcbset{
  promptbox/.style={
    top=10pt,
    colback=lightgray!20,
    colframe=Black,
    colbacktitle=NavyBlue,
    enhanced,
    center,
    attach boxed title to top center={yshift=-0.1in,xshift=0.0in},
    boxed title style={boxrule=0pt,colframe=white,},
  }
}
\newtcolorbox{promptbox}[2][]{promptbox, title=#2,#1}
\tcbset{
  takeawaybox/.style={
    top=10pt,
    colback=lightgray!20,
    colframe=Black,
    colbacktitle=BurntOrange,
    enhanced,
    center,
    attach boxed title to top center={yshift=-0.1in,xshift=0.0in},
    boxed title style={boxrule=0pt,colframe=white,},
  }
}
\newtcolorbox{takeawaybox}[2][]{takeawaybox, title=#2,#1}
\tcbset{
  observationbox/.style={
    top=10pt,
    colback=lightgray!20,
    colframe=Black,
    colbacktitle=YellowGreen,
    enhanced,
    center,
    attach boxed title to top center={yshift=-0.1in,xshift=0.0in},
    boxed title style={boxrule=0pt,colframe=white,},
  }
}
\newtcolorbox{observationbox}[2][]{observationbox, title=#2,#1}

\usepackage{xspace}

\newcommand\blfootnote[1]{%
  \begingroup
  \renewcommand\thefootnote{}\footnote{#1}%
  \addtocounter{footnote}{-1}%
  \endgroup
}

\usepackage{CJK}

\title{How to Set the Batch Size for Large-Scale Pre-training?}

\author[1*$\bigtriangleup$]{Yunhua Zhou}
\author[1,2*$\bigtriangleup$]{Junhao Huang}
\author[1,3]{Shuhao Xing}
\author[1,2]{Yechen Zhang}
\author[1]{Runyu Peng}
\author[1$\dagger$]{Qiping Guo}
\author[3$\dagger$]{Xipeng Qiu}

\affil[1]{Shanghai AI Laboratory}
\affil[2]{Shanghai JiaoTong University}
\affil[3]{Fudan University}

\begin{abstract}
The concept of Critical Batch Size, as pioneered by OpenAI, has long served as a foundational principle for large-scale pre-training. However, with the paradigm shift towards the Warmup-Stable-Decay (WSD) learning rate scheduler, we observe that the original theoretical framework and its underlying mechanisms fail to align with new pre-training dynamics. To bridge this gap between theory and practice, this paper derives a revised $E(S)$ relationship tailored for WSD scheduler, characterizing the trade-off between training data consumption $E$ and steps $S$ during pre-training. Our theoretical analysis reveals two fundamental properties of WSD-based pre-training: 1) $B_{\min}$, the minimum batch size threshold required to achieve a target loss, and 2) $B_{\text{opt}}$, the optimal batch size that maximizes data efficiency by minimizing total tokens. Building upon these properties, we propose a dynamic Batch Size Scheduler. Extensive experiments demonstrate that our revised formula precisely captures the dynamics of large-scale pre-training, and the resulting scheduling strategy significantly enhances both training efficiency and final model quality.

\end{abstract}

\begin{document}

\blfootnote{$\dagger$ Corresponding authors.}
\blfootnote{$*$ Equal contribution. Orders are determined randomly.}
\blfootnote{$\bigtriangleup$ Yunhua Zhou(zhouyunhua@pjlab.org.cn), Junhao Huang(huangjunhao@pjlab.org.cn).}

\maketitle

\section{Introduction}
The continuous evolution of Large Language Models (LLMs) \cite{bi2024deepseek,team2025kimi} is perpetually expanding the frontiers of artificial intelligence, driven the large-scale pre-training. As the scale of pre-training continues to expand, the selection of optimal training strategies becomes paramount. A central challenge in this endeavor is the configuration of batch size to achieve trade-off between training efficiency and performance.

Foundational research on batch size for large-scale pre-training originates from OpenAI \cite{mccandlish2018empirical}, which introduced the concept of \textbf{Critical Batch Size} to characterize the trade-off between token consumption $E$ and steps $S$ during pre-training. Building on this foundation, OpenAI further established the scaling laws for LLMs \cite{kaplan2020scaling}, a milestone that significantly catalyzed the revolution in generative artificial intelligence.

Concurrently, the pre-training paradigm has undergone significant evolution, most notably the transition in learning rate schedulers (LRS). The Warmup-Stable-Decay (WSD) LRS has increasingly replaced the traditional cosine LRS and gained widespread adoption in state-of-the-art models \cite{bi2024deepseek,team2025kimi,hu2024minicpm}. However, we discover that under the WSD LRS, the relationship between data consumption ($E$) and steps ($S$) during pre-training no longer adheres to OpenAI’s original $E(S)$ formula. This discrepancy implies that the underlying mechanism of critical batch size is no longer valid in current pre-training regimes, revealing a profound gap between theoretical foundations and engineering practice.

To bridge this theoretical divide, this paper derives a novel $E(S)$ relationship tailored for modern large-scale pre-training (i.e., adopting WSD LRS). Based on this new formulation, we reveal two intrinsic properties of the ``Stable'' phase in WSD pre-training: \textbf{Threshold Constraint ($B_{\min}$)}: To achieve a specific target loss, the batch size must exceed a certain physical threshold. \textbf{Efficiency Optimality ($B_{\text{opt}}$)}: There exists an optimal batch size that minimizes the total data consumption required to reach the target loss. Furthermore, Based on these insights of both $B_{\min}$ and $B_{\text{opt}}$ exhibit an upward trend as training loss decreases, we introduce a novel batch size scheduler.

The core contributions of this paper are threefold: \\
\textbf{Theoretical Reconstruction:} We are the first to explicitly identify the limitations of existing batch size theories under the WSD paradigm and establish a new $E(S)$ formula that accurately describes the modern pre-training process.\\
\textbf{Property Discovery and Methodological Innovation:} Based on the new $E(S)$ relationship, we reveal two essential properties of the large-scale pre-training—$B_{\min}$ and $B_{\text{opt}}$—and elucidate their evolution mechanisms, leading to a new Batch Size Scheduler for large-scale pre-training.\\
\textbf{Experimental Validation:} Extensive experimental results demonstrate that our proposed $E(S)$ formula precisely captures the dynamics between data consumption ($E$) and steps ($S$) during pre-training, and the resulting Batch Size Scheduler significantly enhances the quality of pre-training.

\section{Related Work}
\subsection{The impact of batch size on model training dynamics}
Batch size, a pivotal hyperparameter in model training, has garnered extensive attention from both academia and industry. \citet{keskar2017large} were among the first to investigate its impact on model generalization, observing that—unlike small batch sizes—training with large batch sizes tends to result in convergence to sharp minima, thereby degrading generalization performance. \citet{mccandlish2018empirical} subsequently introduced a novel perspective by proposing the concept of Critical Batch Size to characterize the trade-off between training efficiency and batch size. Furthermore, they derived the renowned relationship between the total data consumption $E$ and the number of optimization steps $S$ required to reach a specific loss, known as the $E(S)$ formula:
\begin{equation}\label{eq:E(S)}
    (\frac{E}{E_{min}}-1)(\frac{S}{S_{min}}-1)=1.
\end{equation}
Extending the critical batch size framework, \citet{kaplan2020scaling} formalized the scaling laws governing neural language models. They demonstrated that model performance scales as a predictable power-law function of model size, data volume, and compute.

Distinct from Critical Batch Size, Optimal Batch Size characterizes the relationship between batch size and final model performance. However, although scaling laws have driven an exponential increase in model scale, the prohibitive experimental costs have severely limited research into optimal batch size. To address this, \citet{bi2024deepseek} investigated the scaling properties of optimal batch size, revealing that it relates to the compute budget via a power law:
\begin{equation}
    B_{opt}=0.2920·C^{0.3271}.
\end{equation}
Crucially, this scaling law enables the extrapolation of optimal batch sizes for large-scale training from low-cost, small-scale experiments. Beyond compute, recent studies have further established power-law dependencies between optimal batch size and other key dimensions, specifically model size and data volume \cite{shuai2024scaling, li2025predictable}.

While dynamic batch size scheduling was briefly touched upon in large-scale model training~\cite{bi2024deepseek, minimax2025minimax01scalingfoundationmodels}, the theoretical principles guiding these schedules remain undisclosed. This paper aims to bridge this gap by providing a theoretical framework that elucidates the mechanisms underlying these empirical strategies.

\subsection{Scaling relationship between batch size and learning rate}
Given the interdependence of learning rate and batch size, a critical challenge lies in determining the optimal scaling strategy for the learning rate as batch size changes.

\citet{krizhevsky2014one} initially proposed the square-root scaling rule for SGD, suggesting that the learning rate should scale by $\sqrt{k}$ when the batch size scales by $k$. However, this heuristic was subsequently challenged. \citet{goyal2017accurate} demonstrated that for SGD, the learning rate should instead scale linearly with batch size (i.e., by a factor of $k$). \citet{smith2020generalization} corroborated this linear scaling rule, emphasizing its validity specifically within the small-batch regime. Furthermore, while establishing the Critical Batch Size framework, \citet{mccandlish2018empirical} formalized the relationship between optimal learning rate and batch size as follows:
\begin{equation}
    \eta_{opt}=\frac{\eta_{max}}{1+B_{noise}/B},
\end{equation}
where $B_{noise}$ denotes the gradient noise scale, $B$ presents the batch size, and $\eta_{max}$ is a constant. Crucially, in the small-batch regime($B \ll B_{noise}$), the learning rate scales approximately linearly with the batch size.

The widespread adoption of the Adam optimizer \citep{kingma2014adam} has fundamentally altered the relationship between batch size and learning rate.  \citet{you2019large} empirically observed during BERT training that scaling the learning rate by the square root of the batch size ($\eta \propto \sqrt{B}$) yields superior performance. This heuristic was formalized by \citet{liu2019variance}, who demonstrated that under Adam, gradient noise variance scales with $\eta^2/B$; thus, maintaining constant variance requires square-root scaling. \citet{malladi2022sdes} further substantiated this relationship theoretically via a stochastic differential equation (SDE) approximation of Adam. Recently, however, \citet{li2024surge} challenged this convention. By re-examining the optimization dynamics of Adam, they proposed a revised scaling law for the optimal learning rate:
\begin{equation}
    \eta_{opt}=\frac{\eta_{max}}{\frac{1}{2}(\sqrt{\frac{B_{noise}}{B}}+\sqrt{\frac{B}{B_{noise}}})}.
\end{equation}
In the small-batch regime ($B \ll B_{noise}$), the optimal learning rate scales approximately linearly with $\sqrt{B}$. However, once the batch size surpasses the gradient noise $B_{noise}$, the optimal learning rate begins to decay.

\noindent \textbf{Summary and Connection} Prior work falls into two main paradigms: empirically fitting optimal batch size scaling laws (often theory-light) or theoretically deriving learning rate adjustments (often impractical for large-scale training). We address the limitations of both approaches by:

1. Diverging from prior studies that rely exclusively on empirical fitting to determine batch size scaling laws, our work provides a formal theoretical characterization of pre-training dynamics under the WSD schedule. By deriving a novel $E(S)$ relationship for the Stable phase, we establish a robust framework grounded in first principles that elucidates these underlying dynamics;

2. Distinguished from purely theoretical studies on hyperparameters like learning rate and batch size, our work translates theoretical insights into a practical batch size schedule tailored for WSD large-scale pre-training. Validated across diverse scenarios, our approach demonstrates significant practical utility and robustness.

\section{Approach}
\subsection{Rethinking the Critical Batch Size}
To characterize the optimal trade-off between data consumption $E$ and optimization steps $S$, \citet{mccandlish2018empirical} introduced the concept of \textit{Critical Batch Size}. This framework is grounded in the empirical observation that, when training a model to a fixed performance target, $E$ and $S$ satisfy the relationship in Eq.\ref{eq:E(S)}. Here, $S_{min}$ represents the minimum steps required to achieve the target loss, while $E_{min}$ denotes the minimum data volume needed. The Critical Batch Size is formally defined as the ratio $B_{crit} = E_{min}/S_{min}$.

Existing research on Critical Batch Size, including the seminal work by \citet{mccandlish2018empirical}, has predominantly focused on the Cosine learning rate schedule. Crucially, however, the behavior of Critical Batch Size under the Warmup-Stable-Decay (WSD) learning rate schedule \citep{hu2024minicpm} remains significantly underexplored. This represents a critical gap, particularly given the widespread adoption of WSD in modern large-scale pre-training tasks, such as those by DeepSeek \citep{bi2024deepseek}, Kimi \citep{team2025kimi}, and Qwen \citep{yang2025qwen3}.

To analyze this discrepancy, we first reformulate Eq.\ref{eq:E(S)} to examine the data consumption required to reach a specific target loss across varying batch sizes. The reformulated equation is given by:
\begin{equation}\label{E(B)}
    E=E_{min}+BS_{min}.
\end{equation}
This equation indicates that achieving a fixed target loss with a larger batch size typically necessitates greater data consumption. Specifically, assuming a model is trained to the same loss level using batch sizes $B_1$ and $B_2$ (where $B_1 < B_2$), the corresponding data consumption $E_1$ and $E_2$ must satisfy the inequality $E_1 < E_2$.

\begin{figure}
    \centering
    \includegraphics[width=1.0\linewidth]{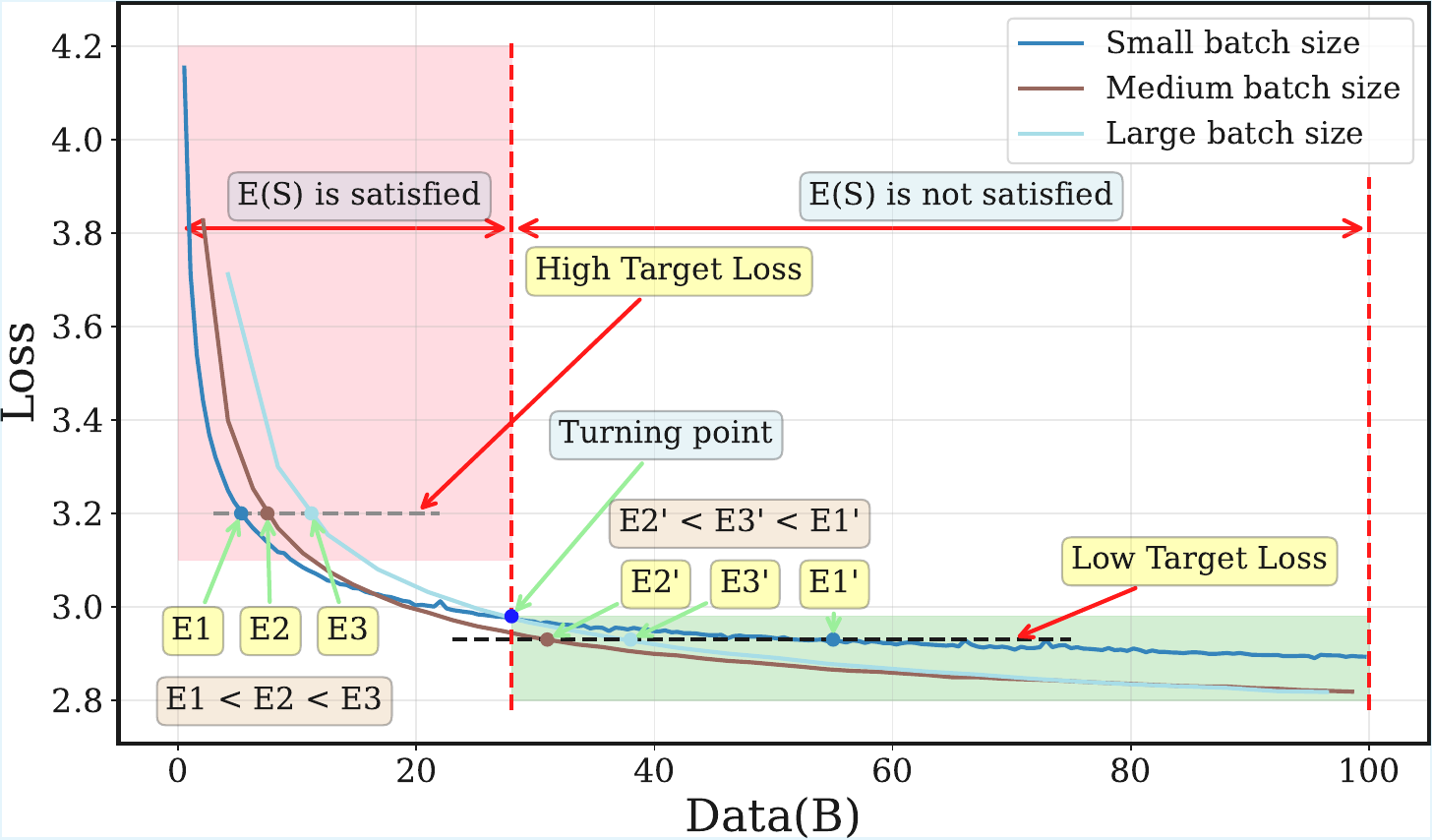}
    \caption{Loss curves for models trained with different batch sizes (Stable phase under WSD schedule). The red region denotes the regime where the $E(S)$ formula and Critical Batch Size theory remain valid. In the green region, the $E(S)$ relationship no longer holds, leading to a failure of the Critical Batch Size framework. Post-intersection, the partial ordering of data consumption among the various batch sizes is inverted.}
    \label{fig:L_S_cross_bs}
\end{figure}

However, under the WSD learning rate schedule, we observe that the training curves $L(D)$ for varying batch sizes intersect during practice. Specifically, while the relationship $E_1 < E_2$ holds at relatively higher target losses, this relationship inverts once the target loss drops below a specific threshold, resulting in $E_1 > E_2$ (as illustrated in Figure \ref{fig:L_S_cross_bs}). This observation stands in direct contradiction to the monotonicity implied by the standard $E(S)$ formula. Consequently, these experimental results demonstrate that the fundamental principles of Critical Batch Size do not hold during the Stable phase of the WSD paradigm.

\subsection{A New E(S) Formula Adapted to Large-scale Pre-training}
Experimental analysis reveals that the prerequisites for the existing Critical Batch Size theory are violated during the Stable phase of the WSD learning rate schedule. Fundamentally, the standard $E(S)$ relationship renders itself inapplicable in this regime. To address this, we draw upon the analytical methodology of \citet{mccandlish2018empirical} regarding SGD optimization dynamics to construct a novel $E(S)$ theoretical framework tailored specifically for the WSD learning rate schedule. This framework models the data consumption $E$ required to reach a target loss as a function of optimization steps $S$, meticulously decomposing the evolution process into three distinct stages:

\noindent \textbf{Initial stage}:$E$ fluctuates inversely with $S-S_{min}$(inverse linear stage in Figure \ref{fig:E(S)});

\noindent \textbf{Transition stage}: $E$ is expressed  as a quadratic function of $S$(transition stage in Figure \ref{fig:E(S)});

\noindent \textbf{Asymptotic stage}: E increases linearly with S(linear stage in Figure \ref{fig:E(S)}).

The corresponding piecewise function expression is as follows:
\begin{equation}\label{E(S)_new}
    E(S) = 
    \begin{cases}
        B_{-1}/(S-S_{min})+B_0, S_{min} <S<S_1, \\
        C(S-S_{opt})^2+E_{min}, S_1<S<S_2, \\
        A_1S+A_0, S>S_2.
    \end{cases}
\end{equation}
For a detailed derivation of this formula, please refer to the Appendix \ref{Optimization_analysis}.

\subsection{Fitting of the New E(S) formula}
From the piecewise form of the function $E(S)$, we obtain 10 parameters to be fitted. First, we impose constraints on these parameters. By requiring the $E(S)$ curve to be continuous, smooth and differentiable, we derive the following equality constraints:
\begin{equation}\label{eqs_1}
    \frac{B_{-1}}{S_1-S_{min}}+B_0=C(S_1-S_{opt})^2+E_{min},
\end{equation}
\begin{equation}\label{eqs_2}
    C(S_2-S_{opt})^2+E_{min}=A_1 S_2+A_0,
\end{equation}
\begin{equation}\label{eqs_3}
    -\frac{B_{-1}}{(S_1-S_{min})^2}=2C(S_1-S_{opt}),
\end{equation}
\begin{equation}\label{eqs_4}
    2C(S_2-S_{opt})=A_1.
\end{equation}
Meanwhile, the following inequality constraints are given:
\begin{equation}\label{ieqs_1}
    S_{min}<S_1<S_{opt}<S_2.
\end{equation}
Thereby, we establish the parameter search space for fitting. The $E(S)$ curve is then fitted by minimizing the Huber loss function \citep{huber1992robust}. Assume the dataset for fitting is $\{(S_i,E_i)\}_{i=1}^{n}$ and the parameters to be fitted are denoted by $\theta$. The fitting process can be described by the following equation:
\begin{equation}
    \theta^*=arg \min_{\theta} \sum_{i=1}^n Huber_{\delta}(E_i,E(S_i,\theta)),
\end{equation}
here, the Huber loss is defined as following:
\begin{equation}
    Huber_{\delta}(x,y)=
    \begin{cases}
        \frac{1}{2}(x-y)^2,|x-y| \le \delta, \\
        \delta|x-y|-\frac{1}{2}\delta^2, |x-y|>\delta.
    \end{cases}
\end{equation}

In order to improve fitting efficiency, we utilize scaling laws to expedite the generation of Loss-Step pairs. According to \citet{luomulti}, in the regime of a constant learning rate, the loss is governed by the following scaling relationship with respect to steps:
\begin{equation}
    L(S)=L_0 +A·S^{-\alpha}.
\end{equation}
Given a target loss, the above formula enables straightforward calculation of the steps needed for the model to descend to that loss. This yields data points for fitting $E(S)$ at the given loss. 

\begin{figure}
    \centering
    \includegraphics[width=1.0\linewidth]{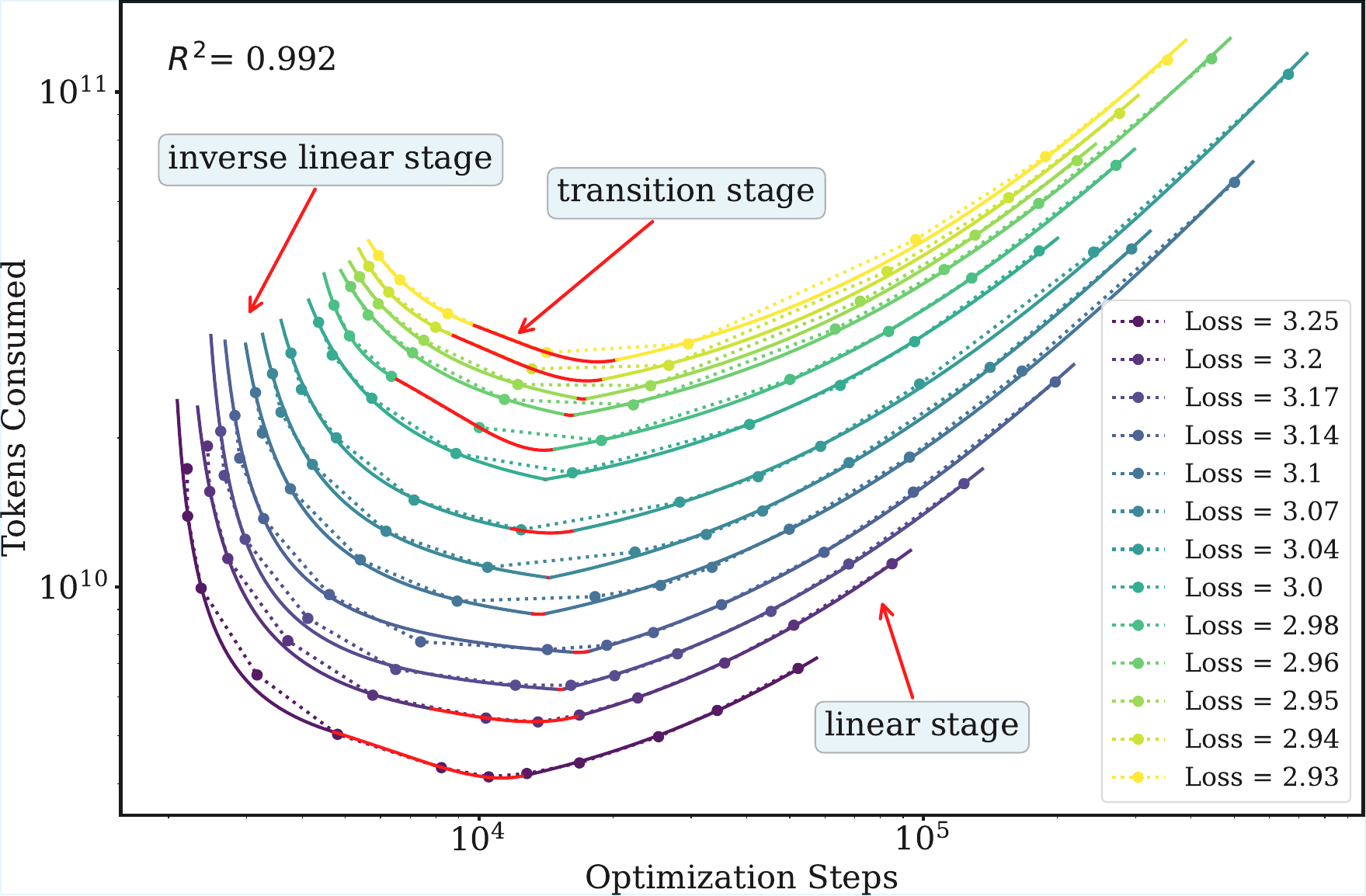}
    \caption{Fitting results of $E(S)$ for 1B model. We select the target loss interval as [2.93, 3.25] and perform fitting on the $E(S)$ curves for target losses within this interval.}
    \label{fig:E(S)}
\end{figure}

Figure \ref{fig:E(S)} presents our new $E(S)$ fitting results for the 1B model. As evident from the plot, the derived $E(S)$ exhibits excellent fitting performance, further substantiating the correctness of our analysis of model training dynamics in the Stable phase.

Under stable learning rate schedule, the Critical Batch Size no longer holds. Instead, it is replaced by two metrics:$B_{min}$ and $B_{opt}$, as given by the following formulas:
\begin{equation}
    B_{min}=A_1,B_{opt}=\frac{E_{min}}{S_{opt}}.
\end{equation}

\begin{figure}
    \centering
    \includegraphics[width=1.0\linewidth]{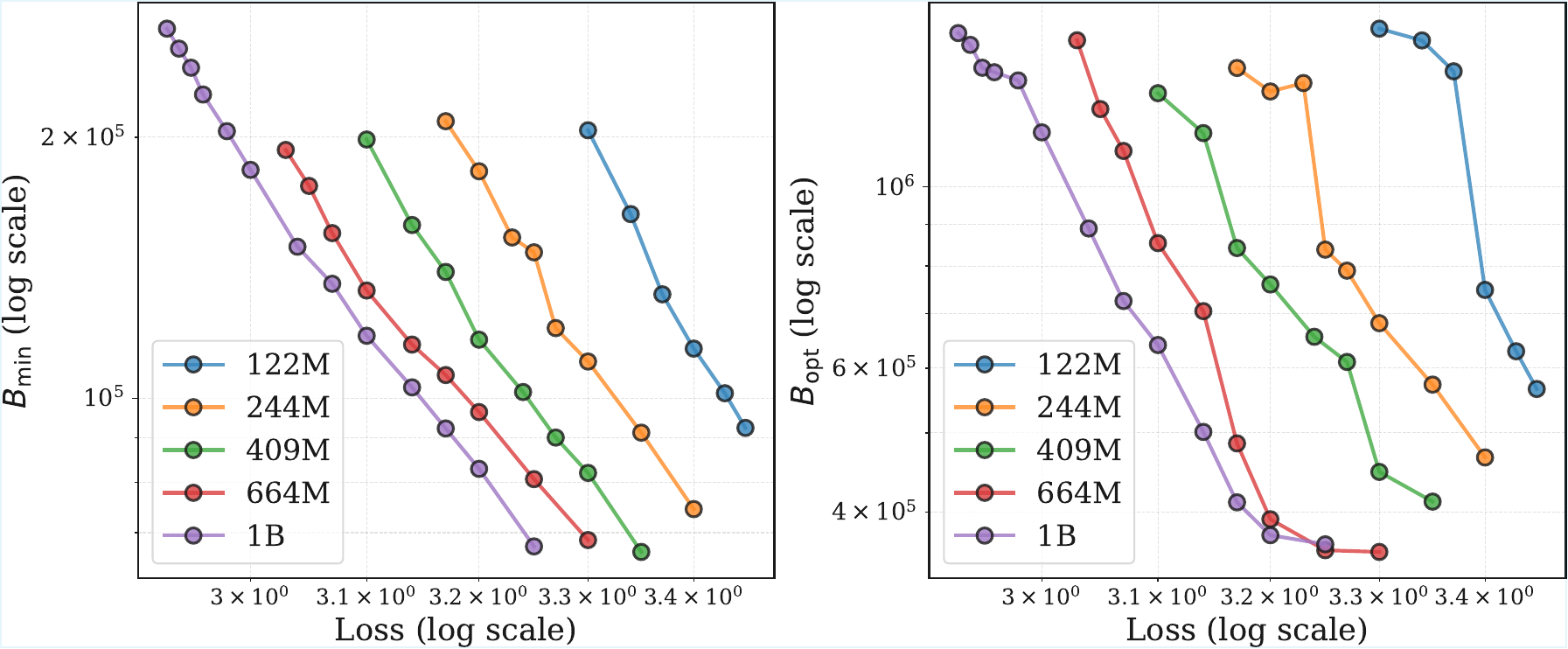}
    \caption{The variation of $B_{min}$ and $B_{opt}$ with respect to loss across different model sizes.}
    \label{fig:Bopt_Bmin}
\end{figure}

Physically, $B_{min}$ and $B_{opt}$ quantify critical batch size thresholds: $B_{min}$ is the minimum batch size needed to reach a target loss, and $B_{opt}$ is the batch size that yields minimum data consumption. Geometrically (see Appendix \ref{Detailed Experiments} for the full $E(S)$ plot), $B_{min}$ equals the slope of the curve's asymptote, while $B_{opt}$ equals the slope from the origin to the curve's minimum. As shown in Figure \ref{fig:Bopt_Bmin}, both metrics scale monotonically with decreasing target loss (increasing data volume). This scaling behavior provides the empirical basis for the dynamic batch size scheduling strategy proposed later.

\subsection{A New Batch Size Schedule}
Within the proposed $E(S)$ theoretical framework, several derived metrics associated with batch size can be established. Specifically, the physical interpretation of $B_{opt}$ is as follows: in the context of constant batch size training, it represents the value that maximizes data efficiency for reaching a specific target loss. Equivalently, it serves as the optimal solution for minimizing loss given a fixed data budget. Nevertheless, employing a static batch size throughout the entire training process is rarely globally optimal in practice. Synthesizing this insight with the experimental observation that $B_{opt}$ increases monotonically as training progresses, we can derive a batch size scheduling scheme better suited for large-scale pre-training. This implies abandoning static batch sizes in favor of a strategy that dynamically expands the batch size over time, thereby achieving superior training performance.

\noindent \textbf{Theorem 1} Assume the model size is fixed, and let the loss be expressed as $L(N,B,D)$, which depends on model size $N$, batch size $B$ and data volume $D$. The two optimization problems below are equivalent:

\noindent \textbf{Problem 1}: For a fixed training data budget, which constant batch size minimizes the model's loss?

\noindent \textbf{Problem 2}: For a prescribed target loss, which constant batch size minimizes the data consumption by the model?

Theorem 1 establishes that modeling the evolution of $B_{opt}$ with respect to data volume is mathematically equivalent to characterizing its relationship with the loss function. While Figure 3 explicitly illustrates the trajectory of $B_{opt}$ as the loss decreases, Theorem 1 implies that this curve simultaneously reveals the scaling law of $B_{opt}$ with respect to data consumption. Given that cumulative data volume serves as a more intuitive metric for training progress than the loss value, we adopt data consumption as the reference benchmark for the dynamic batch size scheduling strategy.

\begin{algorithm} 
	\caption{Batch size scheduling strategy} 
	\label{BS-s} 
	\begin{algorithmic}
		\STATE \textbf{Input:} model size $N$, learning rate $\eta$, the optimal batch size curve $f(N,D)$ under this learning rate, batch size switching interval $D_{interval}$, momentum list $\{\alpha_i\}_{i=1}^{n}$, switching times n.
	    \STATE \textbf{Initialization:} $B_{global,0} = 0,i=1$
        \REPEAT
            \STATE $B_{last}=f(N,(i-1)D_{interval})$
            \STATE $B_{new}=f(N,iD_{inerval})$
            \STATE $B_{global,i}=B_{global,i-1}+(1+\alpha_i)(B_{new}-B_{last})$
            \STATE $i \leftarrow i+1$
        \UNTIL $i>n$
        \STATE \textbf{Output:} $B_{global,1},...,B_{global,n}$
	\end{algorithmic} 
\end{algorithm}
In the subsequent large-scale pre-training experiments, we validate across diverse scenarios that this batch size scheduling strategy effectively enhances model performance.

\section{Experiments}
\subsection{Dataset}
Our experiments utilize the InternLM2 corpus \citep{cai2024internlm2}, categorized into general text, code, and long-context data. The text segment aggregates web pages, academic literature, books, and patents, while the code portion is curated from GitHub and public repositories across languages such as C/C++, Java, and Python. We process the long-context subset via a three-stage pipeline—comprising length selection, statistical filtering, and perplexity-based pruning—to guarantee high-quality long-range dependencies.
\subsection{Model Architectures}
For the $E(S)$ curve fitting experiments, we adopt the InternLM2 architecture. Building upon the LLaMA \citep{touvron2023llama} foundation, InternLM2 fuses the query ($W_q$), key ($W_k$), and value ($W_v$) matrices into a consolidated, interleaved layout per head. Furthermore, the architecture incorporates Grouped-Query Attention (GQA) \citep{ainslie2023gqa} to enhance efficiency.

For the batch size scheduling experiments, we utilize the Qwen3 model series \citep{yang2025qwen3}, comprising both dense and Mixture-of-Experts (MoE) variants. The Qwen3 Dense model refines the Qwen2 architecture \citep{team2024qwen2} by eliminating QKV-bias and incorporating QK-Norm. Meanwhile, the Qwen3 MoE model extends Qwen2.5-MoE by discarding shared experts and adopting a global-batch load balancing loss \citep{qiu2025demons}.

\subsection{Training Settings}
\subsubsection{Fitting of E(S)}
To empirically fit the $E(S)$ curve, we trained five InternLM2 model variants with parameter counts ranging from 122M to 1B, utilizing batch sizes spanning 64K to 7.5M. Optimization was performed using AdamW with a fixed learning rate of $6\times10^{-4}$ and a 1,000-step warmup. The total training volume varied between 50B and 120B tokens depending on the configuration.

\subsubsection{Batch size Scheduling}

\noindent \textbf{Baseline} We conduct our approach using the Qwen3 MoE and Qwen3 Dense architectures. Given the widespread adoption of WSD learning rate schedule in modern large-scale pretraining \citep{team2025kimi,liu2024deepseek,hu2024minicpm}, and acknowledging that the stable phase consumes the majority of the training budget, we focus our experiments on the constant learning rate regime. Specifically, we set the learning rate to $3.2 \times 10^{-4}$ for Qwen3 MoE and $1.75 \times 10^{-4}$ for Qwen3 Dense. For both architectures, we standardize the training configuration with 1,000 warmup steps, a global batch size of 4M, the AdamW optimizer, and a weight decay of 0.1.

\noindent \textbf{Controlled experiments} For comparison, we mirrored the baseline setup while introducing a dynamic batch size strategy. The global batch size was adjusted at 125B-token intervals according to the sequence $\{2\text{M}, 4\text{M}, 5\text{M}, 6\text{M}\}$, achieved by scaling the micro-batch size while keeping other hyperparameters constant.

\subsection{Evaluation}
\subsubsection{Benchmarks}
We evaluate the downstream capabilities of our models using the MMLU benchmark \citep{hendrycks2020measuring} and the CMMLU benchmark \citep{li2024cmmlu}. 
\subsubsection{Evaluation Tools}
For our evaluation, we employ OpenCompass \citep{2023opencompass} to assess model performance on both the MMLU and CMMLU benchmarks. During evaluation, OpenCompass utilizes LMDeploy \citep{2023lmdeploy} to accelerate inference execution.

\subsection{Results}
\begin{figure}
    \centering
    \includegraphics[width=1.0\linewidth]{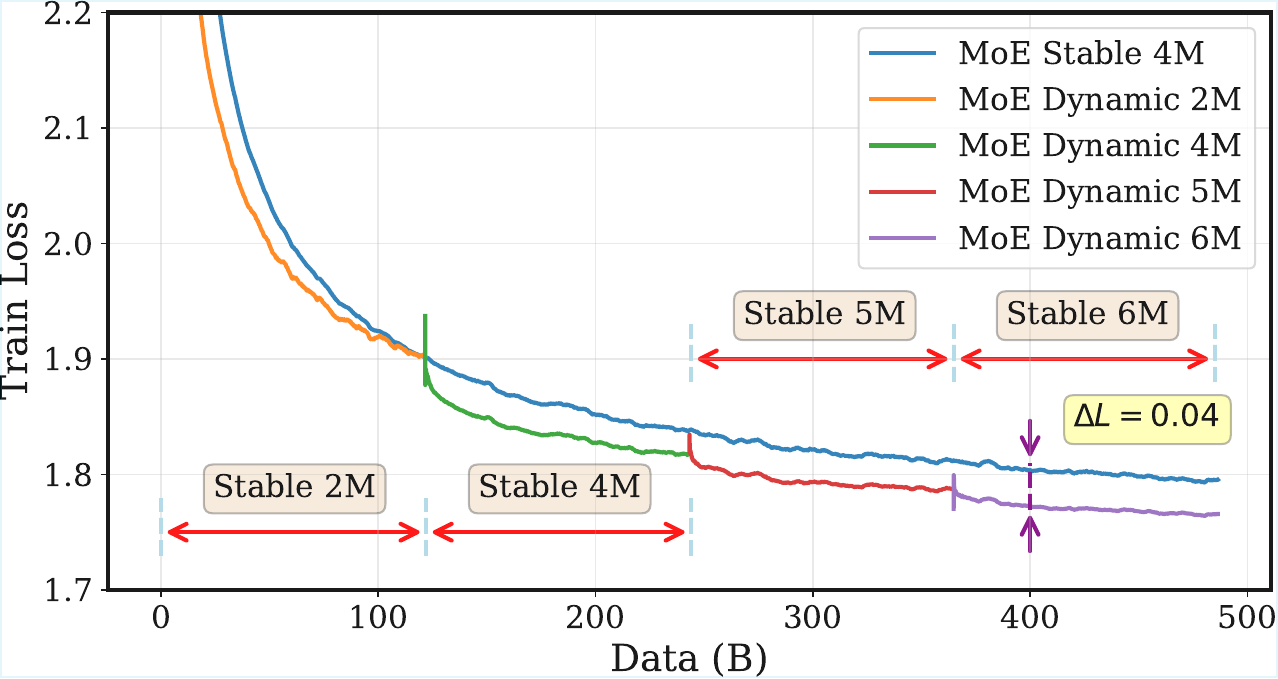}
    \caption{Training loss curves for Qwen3 MoE using fixed and dynamic batch size strategies under a constant learning rate schedule.}
    \label{fig:Ls_MoE_stable}
\end{figure}

\begin{figure}
    \centering
    \includegraphics[width=1.0\linewidth]{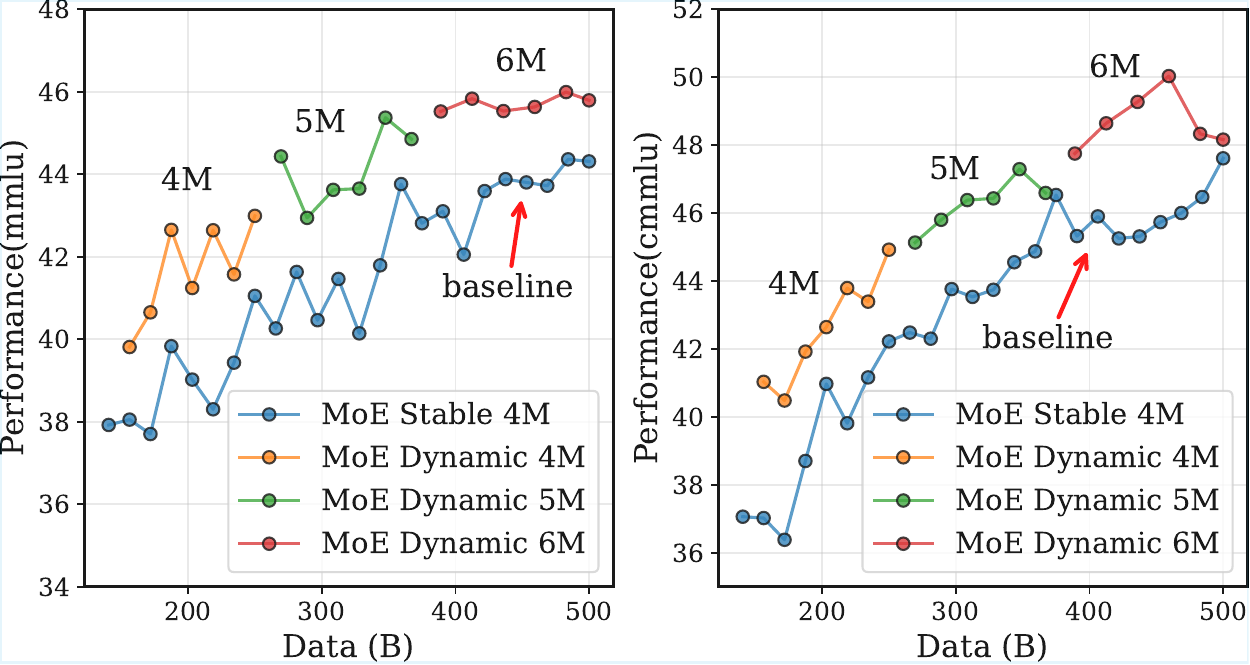}
    \caption{Comparison of downstream benchmark results for Qwen3 MoE under fixed vs. dynamic batch size scheduling at a constant learning rate.}
    \label{fig:Score_MoE_stable}
\end{figure}

\begin{figure}
    \centering
    \includegraphics[width=1.0\linewidth]{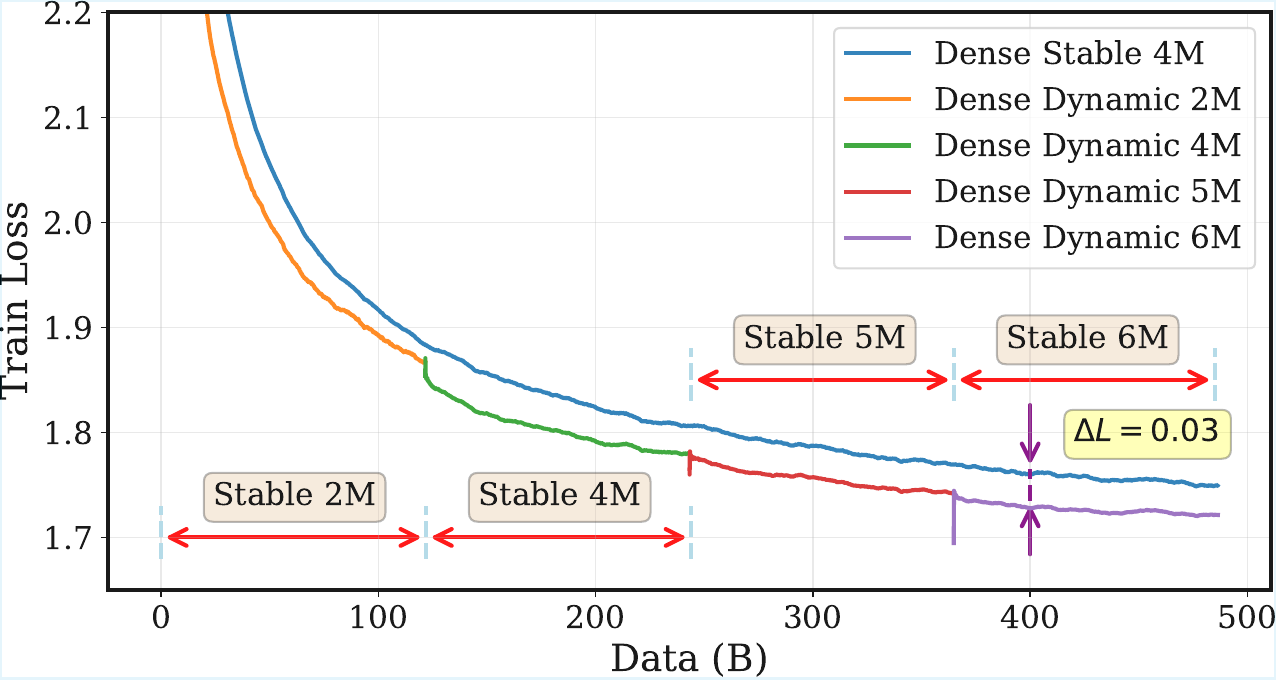}
    \caption{Training loss curves for Qwen3 Dense model under fixed and dynamic batch size strategies at a constant learning rate.}
    \label{fig:Ls_Dense_stable}
\end{figure}

\begin{figure}
    \centering
    \includegraphics[width=1.0\linewidth]{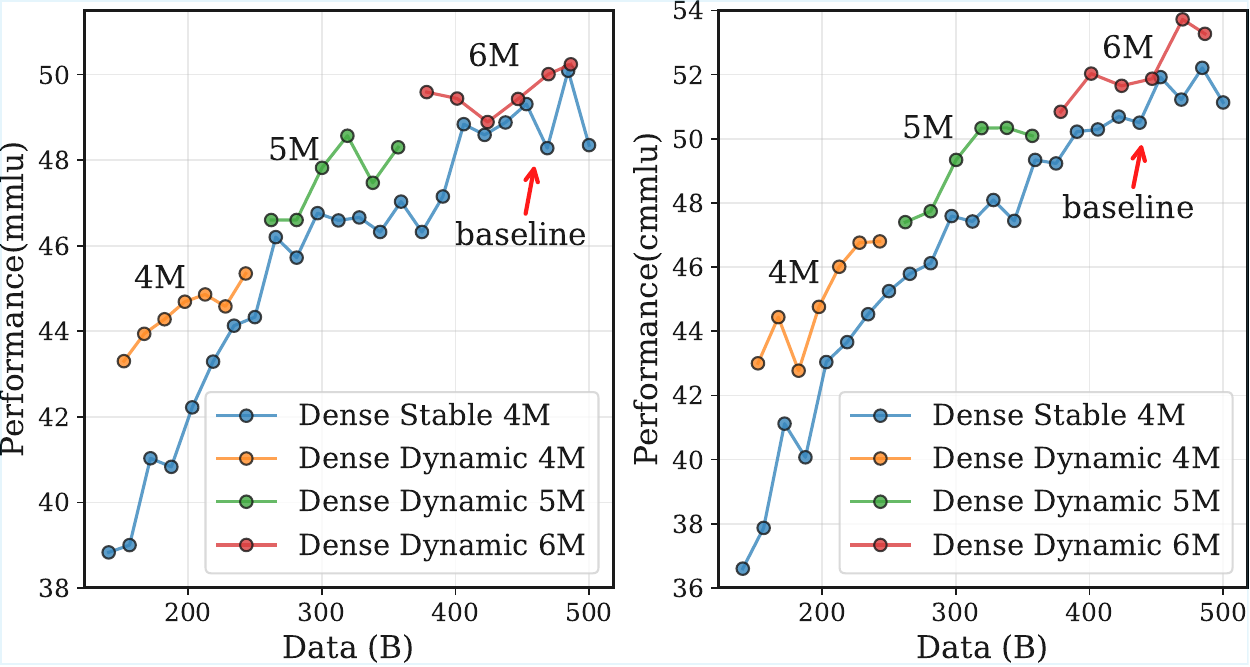}
    \caption{Comparison of downstream benchmark results for Qwen3 Dense under fixed vs. dynamic batch size scheduling at a constant learning rate.}
    \label{fig:score_Dense_stable}
\end{figure}

Figure \ref{fig:Ls_MoE_stable} presents the smoothed training loss trajectory for the Qwen3 MoE model. As illustrated, the curve for the dynamic batch size scheduling strategy consistently lies below that of the fixed-batch baseline, indicating superior convergence. Figure \ref{fig:Score_MoE_stable} further contrasts performance on the MMLU and CMMLU benchmarks, where the dynamic strategy maintains a consistent advantage. Mirroring these findings, Figures \ref{fig:Ls_Dense_stable} and \ref{fig:score_Dense_stable} display the training loss and downstream results for the Qwen3 Dense model, which exhibit identical trends. Collectively, these experiments validate the effectiveness of our dynamic batch size scheduling strategy and corroborate our theoretical analysis of optimization dynamics under WSD learning rate schedule.

\section{Ablations}
\subsection{The Effect of learning rate} 
\begin{figure}
    \centering
    \includegraphics[width=1.0\linewidth]{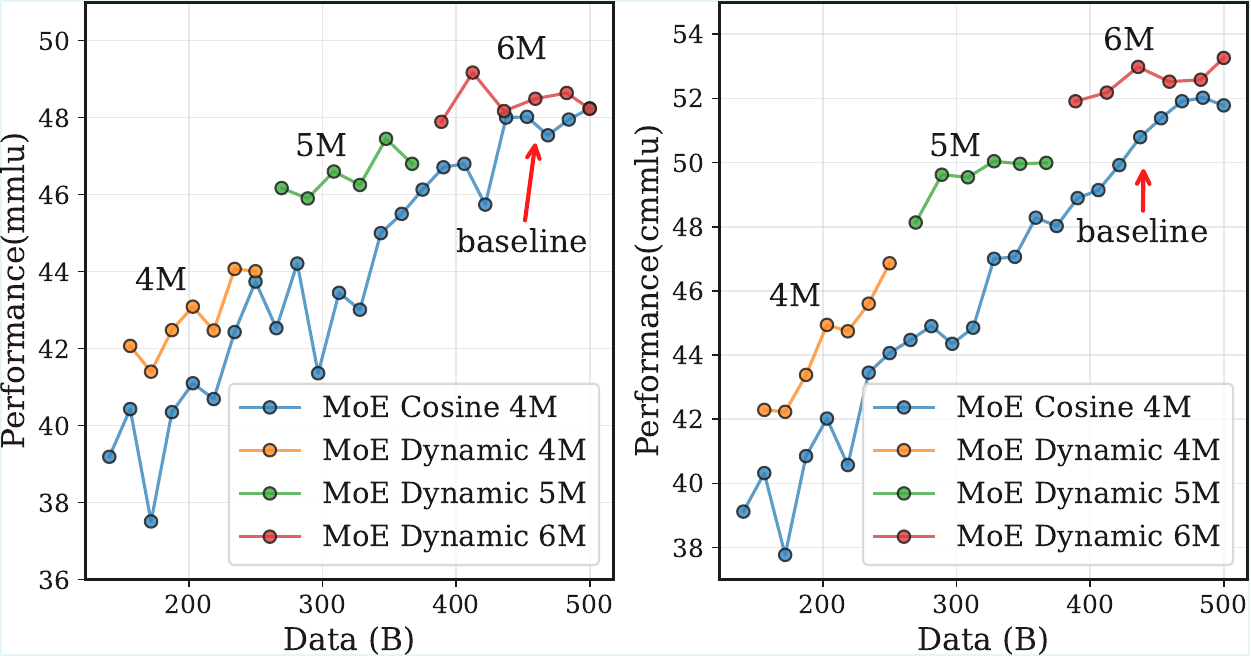}
    \caption{Comparison of downstream benchmark results for Qwen3 MoE under fixed and dynamic batch size scheduling with cosine learning rate schedule.}
    \label{fig:score_MoE_cosine}
\end{figure}

\textbf{Cosine learning rate schedule}
We further validate our strategy's adaptability using a Cosine scheduler on the Qwen3 MoE model (LR: $0 \to 1.7 \times 10^{-3} \to 3.2 \times 10^{-4}$ over 500B tokens). Compared to a fixed 4M batch size baseline, our dynamic schedule—scaling from 2M to 6M in 125B-token increments—yields superior training loss and downstream results (Figure~\ref{fig:score_MoE_cosine}). This success aligns with the Critical Batch Size theory \citep{mccandlish2018empirical}: as gradient noise accumulates during training, expanding the batch size becomes essential to counteract noise-induced instability, thereby facilitating convergence to a deeper loss minimum.

\begin{figure}
    \centering
    \includegraphics[width=1.0\linewidth]{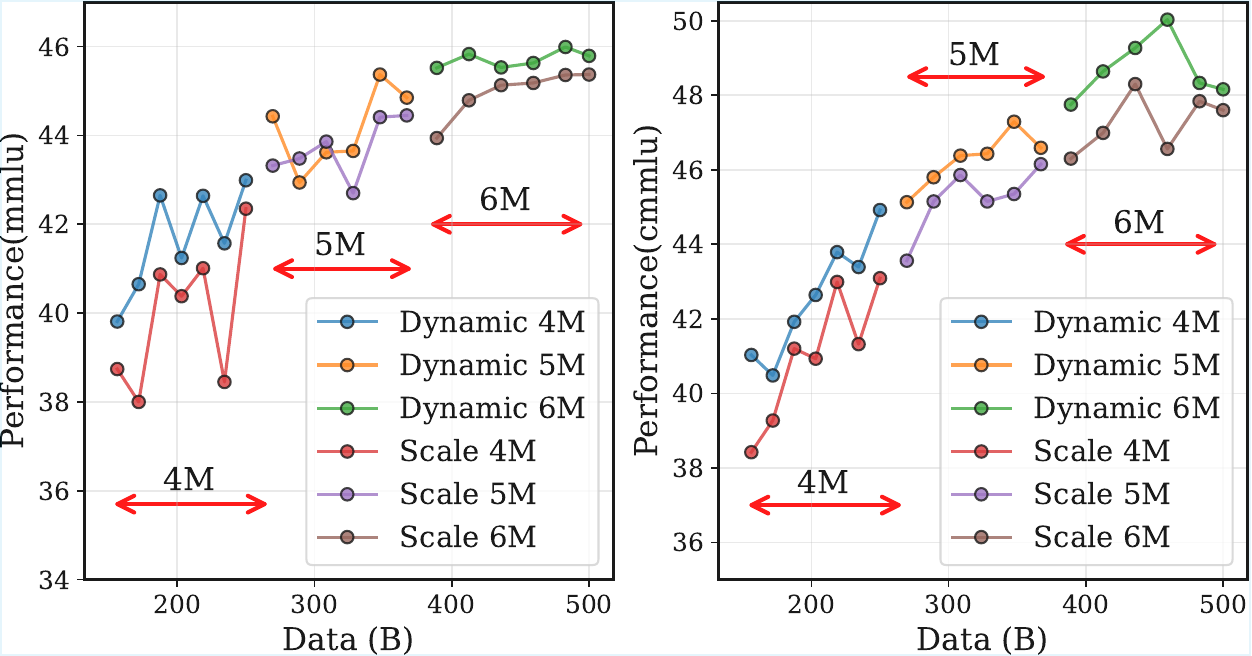}
    \caption{Comparative downstream performance of dynamic batch size scheduling strategies: Constant learning rate versus learning rate scaling regimes.}
    \label{fig:score_MoE_stable_scale}
\end{figure}

\noindent \textbf{Increase the learning rate as batch size increases}
We challenge the convention of scaling the learning rate alongside batch size increases. In an ablation study using the Qwen3 MoE model, we compared square-root scaling ($\text{LR} \propto \sqrt{B}$) against a constant learning rate while progressively increasing the batch size from 2M to 6M. Empirical results in Figure~\ref{fig:score_MoE_stable_scale} demonstrate that scaling the learning rate offers no performance improvement. This is because higher learning rates exacerbate gradient noise, effectively neutralizing the noise-suppression benefits of larger batch sizes and rendering the scaling strategy counterproductive.

\subsection{The Effect of sequence length}
\begin{figure}
    \centering
    \includegraphics[width=1.0\linewidth]{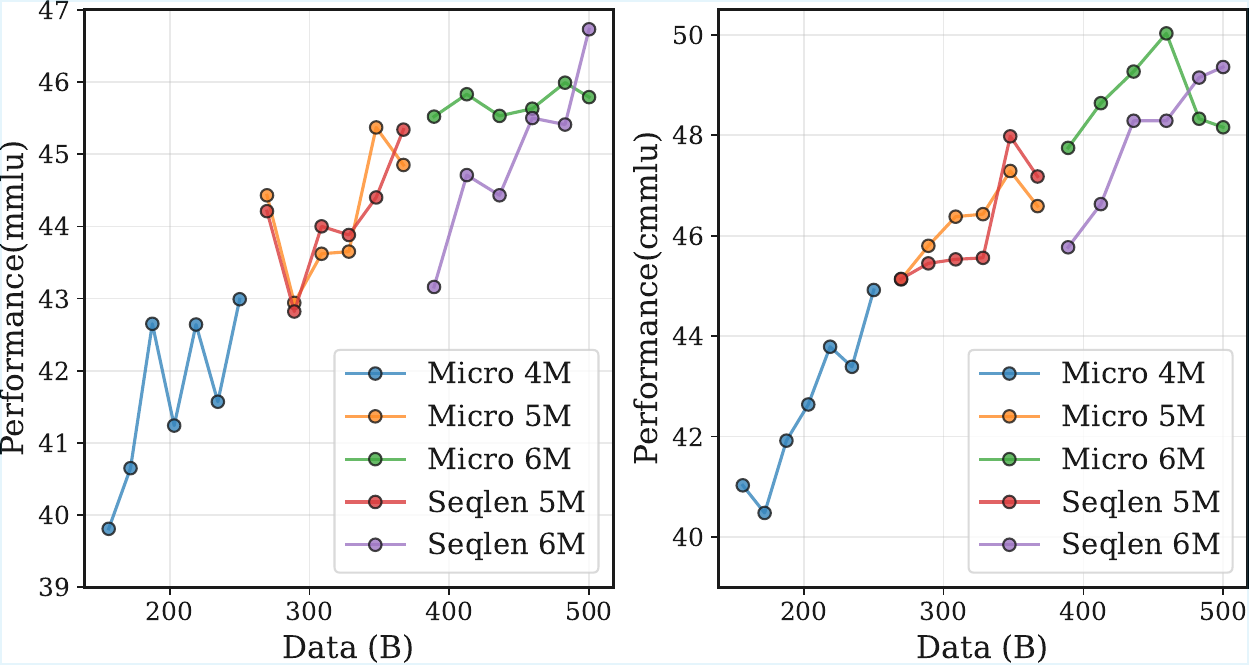}
    \caption{Comparative downstream performance of dynamic batch size scheduling implemented through micro-batch expansion and sequence length extension.}
    \label{fig:score_seqlen}
\end{figure}

An alternative to micro-batch scaling is the extension of sequence length (\textit{seqlen}) to achieve larger global batch sizes. We evaluated this approach on Qwen3 MoE, comparing a 4K \textit{seqlen} baseline against a strategy that shifted \textit{seqlen} to 5K and 6K at specific intervals (250B and 375B tokens). While both methods reach equivalent batch sizes, \textit{seqlen} extension perturbs the training distribution by altering the sample structure. Empirical results in Figure~\ref{fig:score_seqlen} reveal an immediate performance drop upon increasing \textit{seqlen} to 6K, suggesting a non-trivial adaptation period is necessary to reconcile the distribution shift. Although the model eventually recovers, the risk of a learning preference shift toward long-context sequences makes this approach less desirable for standard large-scale pre-training. 
\subsection{The Effect of weight decay}
\begin{figure}
    \centering
    \includegraphics[width=1.0\linewidth]{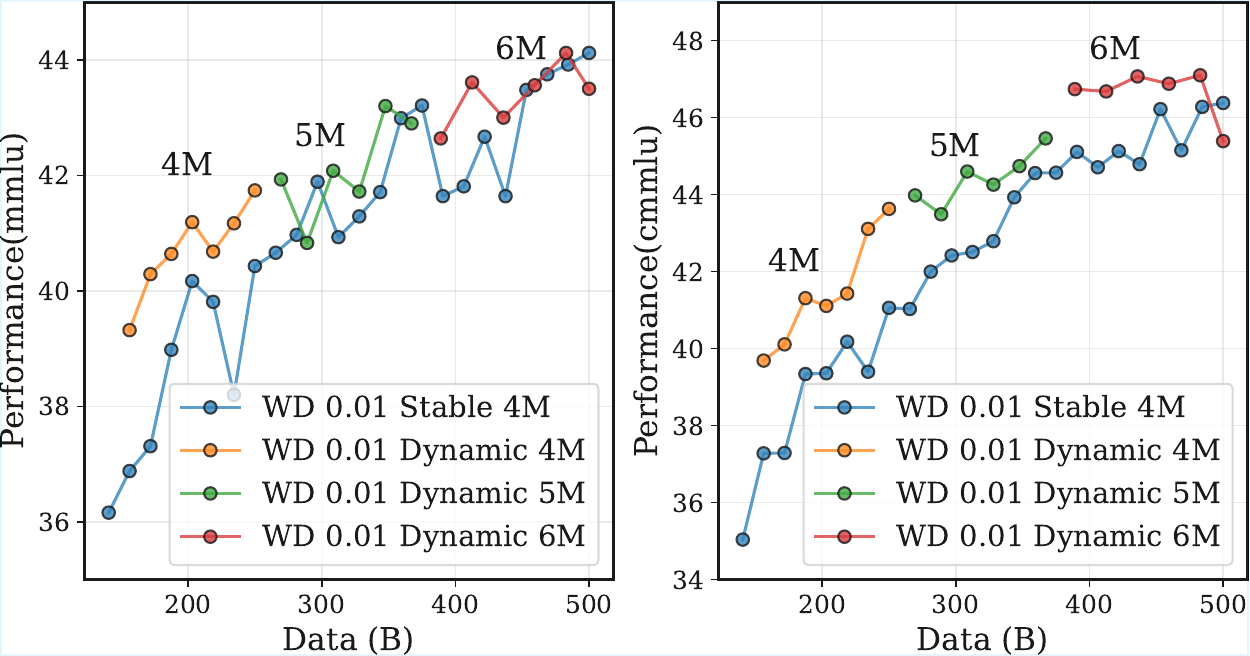}
    \caption{Comparative downstream performance of fixed and dynamic batch size scheduling under different weight decay settings.}
    \label{fig:score_stable_wd}
\end{figure}

We further investigate the sensitivity of our strategy to weight decay. By reducing the coefficient to 0.01 on the Qwen3 MoE model, we observe in Figure \ref{fig:score_stable_wd} that the initial advantage of the dynamic strategy diminishes and nearly vanishes as training progresses. Furthermore, a cross-comparison with the main experiments (Figure \ref{fig:Score_MoE_stable}, WD=0.1) confirms that the 0.01 setting results in significantly inferior baselines. These findings indicate that the effectiveness of dynamic batch sizing is coupled with regularization strength; specifically, the full benefits of the strategy are contingent upon an optimal weight decay setting.

\subsection{The Effect of Continued Training}
\begin{figure*}[t]
    \centering
    \includegraphics[width=1.0\linewidth]{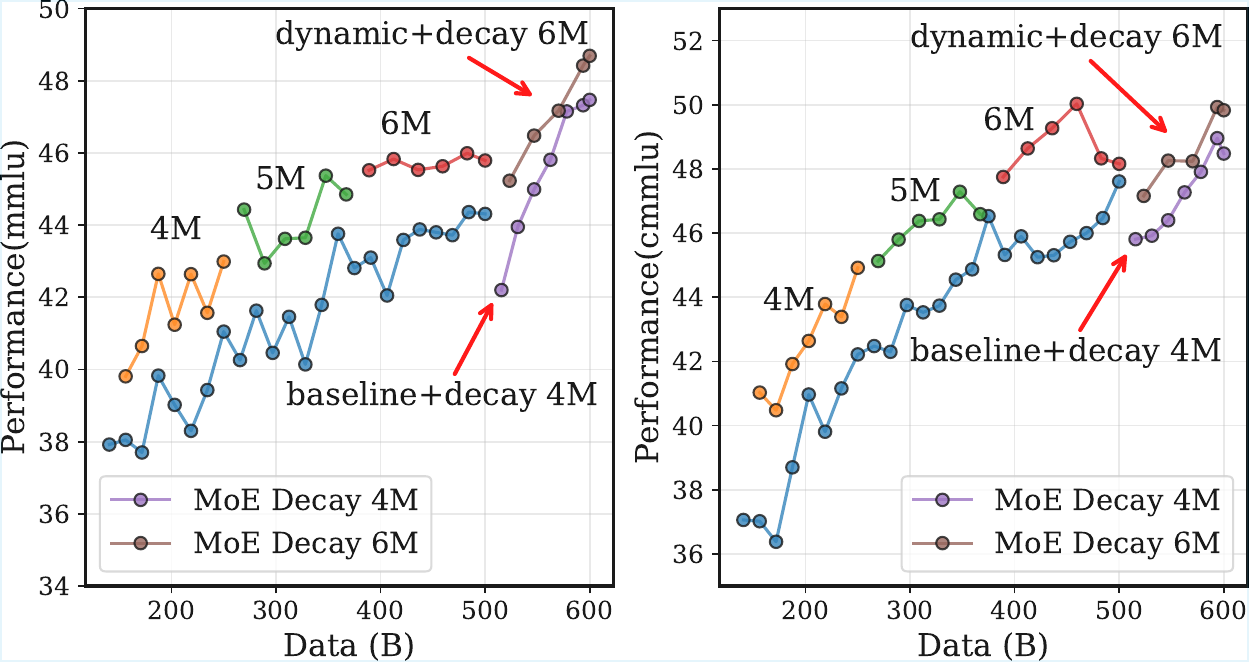}
    \caption{Comparison of downstream benchmark results for fixed and dynamic batch size strategies in the annealing phase.}
    \label{fig:score_MoE_stable_decay}
\end{figure*}

To validate compatibility with modern pretraining protocols, we extended our evaluation to the decay phase of the WSD schedule, characterized by high-quality data annealing. Using the pre-trained MoE model, we conducted a comparative run over 100B tokens with a linear learning rate decay to 10\%. The baseline maintained a 4M batch size, whereas the dynamic strategy retained its peak 6M batch size. Figure \ref{fig:score_MoE_stable_decay} demonstrates that the performance advantage of the dynamic strategy is sustained throughout this phase. This confirms the robustness of our approach against data distribution shifts, validating its effectiveness in standard large-scale training pipelines.

\section{Conclusion}
This work first elucidates the limitations of the seminal Critical Batch Size theory \citep{mccandlish2018empirical}, demonstrating its inapplicability to the Warmup-Stable-Decay (WSD) scheduler prevalent in modern large-scale pre-training. To address this gap, we propose a novel $E(S)$ formulation tailored specifically for the WSD paradigm. Within this framework, we identify two pivotal metrics: $B_{min}$, the minimum batch size threshold required to reach a target loss, and $B_{opt}$, the optimal batch size for maximizing data efficiency. We observe that both $B_{min}$ and $B_{opt}$ increase monotonically as training loss decreases. Motivated by this finding, we introduce a dynamic batch size adjustment strategy and validate its effectiveness across multiple large-scale pre-training scenarios.

\section*{Limitations}
While our batch size adjustment paradigm proves effective for large-scale pre-training, it is circumscribed by certain limitations. (1) Computational costs restricted our $E(S)$ curve fitting to a specific learning rate ($6 \times 10^{-4}$), leaving its behavior under other learning rates unexplored. (2) Although empirically successful, the dynamic strategy has not yet been formalized with a theoretical proof. (3) The training instabilities associated with sequence length (\textit{seqlen}) switching remain unaddressed, limiting the overall flexibility of the scheduling strategy. We aim to explore these aspects in subsequent studies.
\section*{Use of AI Assistants}

We primarily use AI assistants to improve and enrich our writing.

\clearpage
\bibliographystyle{plainnat}
\bibliography{refs}


\clearpage
\appendix

\section{Appendix}
\subsection{Proof of the theorem}
\noindent \textbf{Theorem 1} Assume the model size is fixed, and let the loss be expressed as $L(N,B,D)$, which depends on model size $N$, batch size $B$ and data volume $D$. The two optimization problems below are equivalent:

\noindent \textbf{Problem 1}: For a fixed training data budget, which constant batch size minimizes the model's loss?

\noindent \textbf{Problem 2}: For a prescribed target loss, which constant batch size minimizes the data consumption by the model?

\noindent \textbf{Proof} We first express the two problems in the framework of optimization theory.

\noindent \textbf{Problem 1}
\begin{equation}
\begin{split}
    \min_{B} L(N,&B,D) \\
    s.t. D=&D_0.
\end{split}
\end{equation}

\noindent \textbf{Problem 2}
\begin{equation}
\begin{split}
    \min_{B} &D \\
    s.t. L(N,B&,D)=L_0.
\end{split}
\end{equation}
Let $D_0$ be fixed, and define:
\begin{equation}
    L_0=\min_{B}L(N,B,D_0)=L(N,B^*,D_0).
\end{equation}
Taking $L_0$ as the target loss in Problem 2, we prove that the solution to Problem 2 is also $B^*$. To do so, it suffices to show that training with batch size $B^*$ consumes less data than any other batch size. 
\begin{equation}
    \forall B,L(N,B,D)=L_0=L(N,B^*,D_0) \leq L(N,B,D_0).
\end{equation}
Since $L(N,B,D)$ is monotonically decreasing in $D$, it necessarily follows that:
\begin{equation}
    D_0 \leq D.
\end{equation}
Thus, the solution to Problem 2 is also $B^*$. Problem 1 and Problem 2 are equivalent. Q.E.D.

\subsection{Analysis of the Model Optimization Process in the Stable Phase}\label{Optimization_analysis}
\subsubsection{Analysis of Training Dynamics}
Following the analysis of the model training process under SGD optimization by \citet{mccandlish2018empirical}, we re-analyzed the training dynamics under the condition of stable learning rate schedule. Using the Taylor expansion formula, we performed a quadratic approximation of the loss curve, yielding:
\begin{equation}\label{Taylor}
    L(\theta-\epsilon V) \approx L(\theta)-\epsilon G^TV+\frac{1}{2}\epsilon^2V^THV,
\end{equation}
where $\theta$ is model parameter, $G$ is gradient, $H$ is matrix of Hessian, $V$ is descending direction, $\epsilon$ is learning rate. Let $B$ denotes the batch size in SGD. The stochastic gradient estimate at each step takes the following form:
\begin{equation}
    G_{est}(\theta)=\frac{1}{B} \sum_{i=1}^{B} G_i.
\end{equation}
We assume that $G_i$ are i.i.d, $G_i \sim N(G, \Sigma)$, then we have:
\begin{equation}
    E[G_{est}(\theta)]=G,Cov(G_{est}(\theta))=\frac{\Sigma}{B}.
\end{equation}
We substitute $V=G_{est}(\theta)$ into formula (\ref{Taylor}). Taking the expectation of both sides, we obtain:
\begin{equation}
    E[L(\theta-\epsilon G_{est})]=L(\theta)-\epsilon|G|^2+\frac{1}{2}\epsilon^2(G^THG+\frac{tr(H\Sigma)}{B}).
\end{equation}
Hence,
\begin{equation}\label{E(L)}
E[\Delta L]=E[L(\theta)-L(\theta-\epsilon G_{est})] =\epsilon|G|^2-\frac{1}{2}\epsilon^2(G^THG+\frac{tr(H\Sigma)}{B}).
\end{equation}
For analytical convenience, we approximate the Hessian matrix as the identity matrix.Then, the formula (\ref{E(L)}) can be approximated as:
\begin{equation}\label{E(L)_approximate}
    E[\Delta L] \approx \epsilon|G|^2-\frac{1}{2}\epsilon^2(|G|^2+\frac{tr(\Sigma)}{B}) .
\end{equation}
Using formula (\ref{E(L)_approximate}) as our foundation, we investigate how the per-step loss reduction varies with different batch sizes. For full-batch gradient descent, we have:
\begin{equation}
    E[\Delta L]_{full-batch}=\epsilon|G|^2-\frac{1}{2}\epsilon^2|G|^2.
\end{equation}
To achieve the same amount of loss reduction as one full-batch gradient descent step, the required number of steps for batch size $B$ is given by:
\begin{equation}\label{delta_S}
    \delta S=\frac{E[\Delta L]_{full-batch}}{E[\Delta L]_B}=\frac{1-\frac{1}{2}\epsilon}{1-\frac{1}{2}\epsilon(1+\frac{B_{noise}}{B})},
\end{equation}
where $B_{noise}=tr(H)/|G|^2$ denotes the gradient noise scale of the model. Since the learning rate is very small in practice, we can approximate formula ($\ref{delta_S}$) as:
\begin{equation}
    \delta S \approx \frac{1}{1-\frac{1}{2}\epsilon \frac{B_{noise}}{B}}.
\end{equation}
The volume of training data processed by the model thus far is:
\begin{equation}
    \delta E=B\delta S \approx \frac{B}{1-\frac{1}{2}\epsilon \frac{B_{noise}}{B}}.
\end{equation}
Given that the model reaches loss $L$ after $S_{min}$ steps under full-batch gradient descent, the required step $S$ and data volume $E$ to achieve the same loss with batch size $B$ are respectively:
\begin{equation}
    S=\int_0^{S_{min}} \delta S ds \approx \int_0^{S_{min}}\frac{1}{1-\frac{1}{2}\epsilon \frac{B_{noise}(s)}{B}} ds,
\end{equation}
\begin{equation}
    E=\int_0^{S_{min}}\delta Eds \approx \int_0^{S_{min}}\frac{B}{1-\frac{1}{2}\epsilon \frac{B_{noise}(s)}{B}} ds.
\end{equation}

\subsubsection{Asymptotic Analysis}
We analyze the asymptotic behavior of $E(S)$ curve at both ends:as $S \rightarrow S_{min}$ and as $S \rightarrow +\infty$.

\noindent \textbf{1.{$S \rightarrow S_{min}$}}

\noindent When $S$ approaches $S_{min}$, it corresponds to the case where the batch size tends to infinity. We analyze the scaling relationship captured by the product $(S-S_{min})E$:
\begin{equation}
    (S-S_{min})E = \int_0^{S_{min}}\frac{B·\frac{1}{2}\epsilon B_{noise}(s)}{B-\frac{1}{2}\epsilon B_{noise}(s)}ds\int_0^{S_{min}}\frac{B}{B-\frac{1}{2}\epsilon B_{noise}(s)}ds.
\end{equation}
Letting $B \rightarrow +\infty$, we have:
\begin{equation}
    \lim_{B \to +\infty}(S-S_{min})E=S_{min}·\int_0^{s_{min}} \frac{1}{2}\epsilon B_{noise}(s)ds.
\end{equation}
Using an infinite series expansion, we can express $E(S)$ as:
\begin{equation}
    E(S)=\frac{B_{-1}}{S-S_{min}}+ \sum_{i=0}^{\infty} B_i(S-S_{min})^i.
\end{equation}
By truncating the higher-order terms, the above formula is approximated as:
\begin{equation}
    E(S) \approx \frac{B_{-1}}{S-S_{min}}+B_0.
\end{equation}
\textbf{2.{$S \rightarrow +\infty$}}

\noindent We note that, since the learning rate is constant, the necessary and sufficient condition for the loss curve to continue decreasing under batch size $B$ is:
\begin{equation}
\begin{split}
    E[\Delta L]_B > 0 \Leftrightarrow 1 - \frac{1}{2}\epsilon \frac{B_{noise}}{B}>0\Leftrightarrow B>\frac{1}{2}\epsilon B_{noise}.
\end{split}
\end{equation}
In other words, to sustain loss reduction, the batch size must be larger than a dynamic lower bound that scales with the instantaneous gradient noise. Thus, when loss stagnation occurs, the batch size has effectively hit this bound, signaling convergence. Formally, we have:
\begin{equation}\label{E_S_infty}
    \lim_{S \to +\infty}\frac{E}{S}=A_1.
\end{equation}
Similarly, formula (\ref{E_S_infty}) can also be expressed in the form of an infinite series:
\begin{equation}
    E(S)=A_1S + \sum_{i=-\infty}^{0}A_iS^i.
\end{equation}
By truncating the higher-order terms, the above formula is approximated as:
\begin{equation}
    E(S) \approx A_1S+A_0.
\end{equation}
\subsubsection{Reconstruction of E(S)}
Through above asymptotic analysis of $E(S)$ curve, we have understood the forms that $E(S)$ takes when $S$ tends to $S_{min}$ and to infinity, respectively. What remains an open question is the variation of $E(S)$ when $S$ falls within the intermediate interval. Since $E(S) \rightarrow +\infty$ as $S \rightarrow S_{min}$ and as $S \rightarrow +\infty$, Rolle's Theorem implies the existence of a point $S^* \in (S_{min}, +\infty)$ such that $E'(S^*)=0$. That is, $E(S)$ has a minimum point, at which the model reaches data optimality - consuming the least amount of data.

Lacking a tractable closed-form expression for $E(S)$ in the intermediate regime, we approximate the curve with a piecewise function. The specific expression for $E(S)$ can be found in formula (\ref{E(S)_new}).Meanwhile, we require that $E(S)$ be continuous, smooth, and differentiable, thereby leading to equality constraint conditions, from formula (\ref{eqs_1}) to formula (\ref{eqs_4}). simultaneously, we require that $E(S)$ has an extreme point the quadratic function stage, thus an inequality constraint is imposed as given in formula (\ref{ieqs_1}).

\subsection{Detailed Experimental Settings and Results}\label{Detailed Experiments}
\subsubsection{Fitting of the New E(S) Formula}
For the empirical fitting of our proposed $E(S)$ formulation, we employ the InternLM2 architecture, training 5 model variants across 13 distinct batch size configurations. The architectural specifications for these models are summarized in Table~\ref{tab:model_configs}, while their corresponding batch size experimental setups are detailed in Table~\ref{tab:batch_size_configs}.

\begin{figure}
    \centering
    \includegraphics[width=1.0\linewidth]{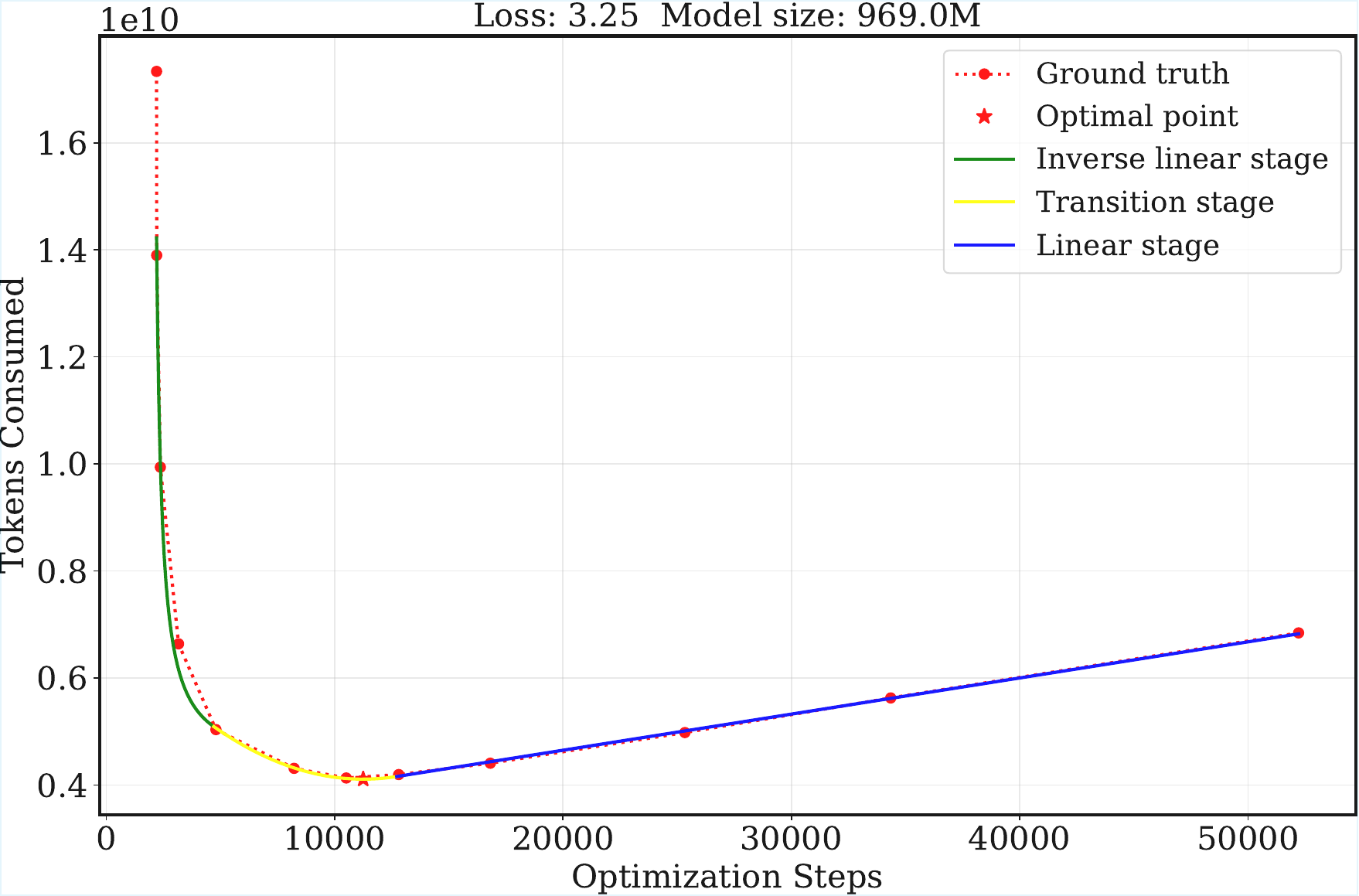}
    \caption{For a fixed loss of 3.25, the $E(S)$ curve of the InternLM2-1B model demonstrates a tripartite structure. Specifically, the optimization process is partitioned into three functional stages: the \textit{Inverse Linear Stage}, the \textit{Transition Stage}, and the \textit{Linear Stage}, each representing a different scaling relationship between data consumption and training steps.}
    \label{fig:loss=3.25}
\end{figure}

Since the $E(S)$ curves in Figure \ref{fig:E(S)} are presented in log-log space, intuiting their progression in a linear coordinate system can be challenging. To provide a clearer physical interpretation, we select a fixed loss threshold and illustrate the corresponding $E(S)$ relationship in linear coordinates in Figure \ref{fig:loss=3.25}.

\begin{table}
\centering
\caption{Architectural configurations of the InternLM2 model series.}
\label{tab:model_configs}
\begin{tabular}{|l|c|c|c|c|}
\hline
\textbf{Models} & \textbf{Hidden Size} & \textbf{Layers} & \textbf{Heads (KV/Q)} & \textbf{MLP Ratio} \\ \hline
InternLM2-122M  & 1024                 & 12              & 2/32                  & 2.5                \\ \hline
InternLM2-244M  & 1280                 & 15              & 2/32                  & 2.5                \\ \hline
InternLM2-409M  & 1536                 & 18              & 2/32                  & 2.5                \\ \hline
InternLM2-664M  & 1792                 & 21              & 2/32                  & 2.5                \\ \hline
InternLM2-1B    & 2048                 & 24              & 2/32                  & 2.5                \\ \hline
\end{tabular}
\end{table}

\begin{table}
\centering
\caption{Batch size configurations for different InternLM2 model scales in the $E(S)$ fitting experiments.}
\label{tab:batch_size_configs}
\begin{tabular}{|l|p{12cm}|} 
\hline
\textbf{Models} & \textbf{Batch Sizes} \\ \hline
InternLM2-121M  & 128k, 256k, 512k, 1M, 2M, 4M, 6M, 7.5M \\ \hline
InternLM2-244M  & 128k, 256k, 512k, 1M, 2M, 4M, 6M, 7.5M \\ \hline
InternLM2-409M  & 128k, 256k, 512k, 1M, 2M, 4M, 6M, 7.5M \\ \hline
InternLM2-664M  & 128k, 256k, 512k, 1M, 2M, 4M, 6M, 7.5M \\ \hline
InternLM2-1B    & 64k, 128k, 160k, 192k, 256k, 320k, 384k, 512k, 1M, 2M, 4M, 6M, 7.5M \\ \hline
\end{tabular}
\end{table}

\subsubsection{Experimental Settings and Results of Ablations}
The architectural configurations for the Qwen3 models are as follows. The Qwen3-Dense model features a hidden dimension of 2,048, 48 layers, and an attention mechanism with 32 query heads and 4 key-value (KV) heads, totaling approximately 2 billion (B) parameters. The Qwen3-MoE model is configured with a hidden dimension of 1,024, 24 layers, 32 query heads and 4 KV-heads. This Mixture-of-Experts (MoE) variant incorporates 128 total experts, with 8 experts activated per token. While the total parameter count for the MoE model is 4B, it maintains only 538M active parameters per forward pass.

\begin{figure}
    \centering
    \includegraphics[width=1.0\linewidth]{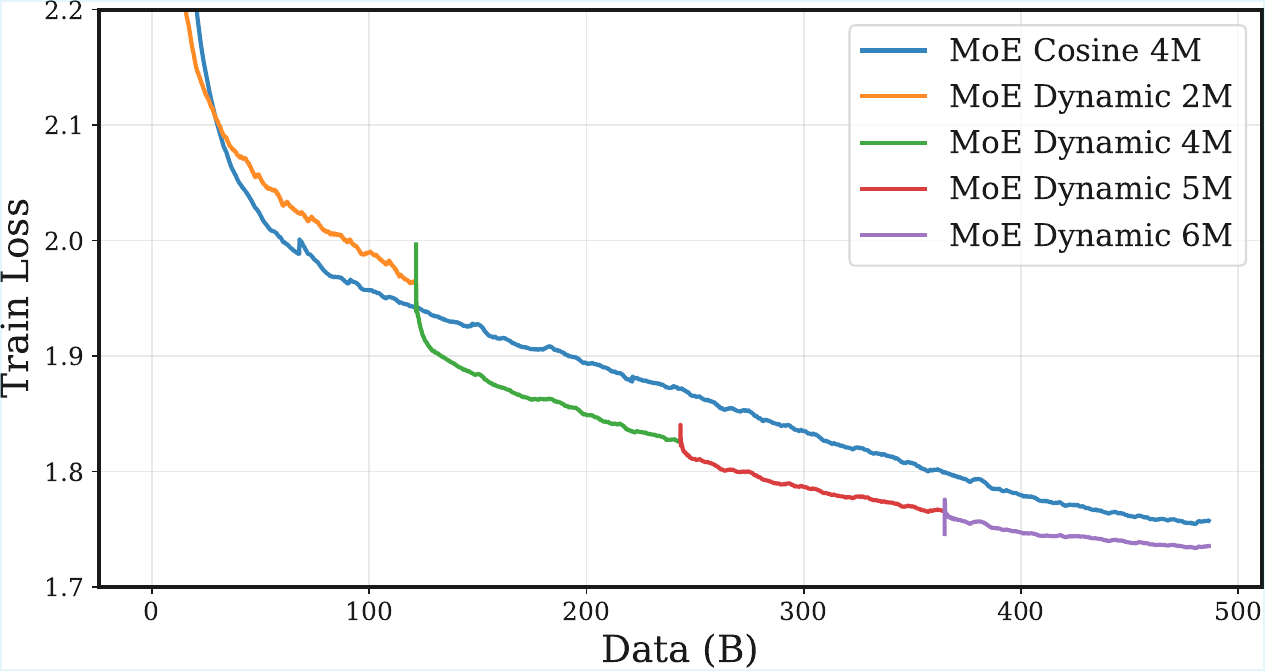}
    \caption{Loss curves of the Qwen3 MoE model trained with fixed and dynamic batch size strategies under cosine learning rate schedule.}
    \label{fig:loss_MoE_cosine}
\end{figure}

\noindent \textbf{Cosine learning rate schedule}
The learning rate follows a cosine schedule, which linearly warms up from 0 to $1.7 \times 10^{-3}$ over the first 1,000 steps, followed by a cosine decay to $3.2 \times 10^{-4}$ over a total training duration of 500B tokens. For the baseline configuration, we maintain a constant batch size of 4M. In contrast, our dynamic batch size strategy progressively scales the batch size at intervals of 125B tokens, following the sequence: 2M, 4M, 5M, and 6M. The training result is shown in Figure \ref{fig:loss_MoE_cosine}.

\begin{figure}
    \centering
    \includegraphics[width=1.0\linewidth]{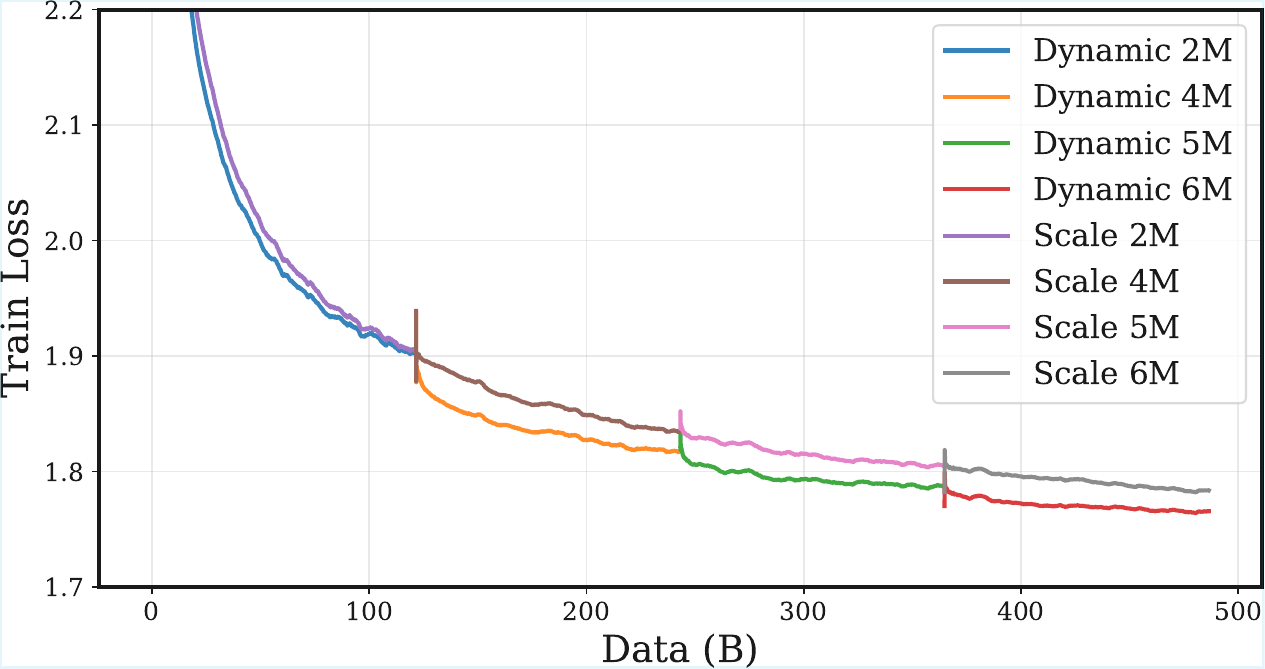}
    \caption{Comparison of training loss trajectories for dynamic batch size strategies featuring constant LR versus LR scaling proportional to the batch size.}
    \label{fig:loss_stable_scale}
\end{figure}

\noindent \textbf{Increase the learning rate as batch size increases}
Using the Qwen3 MoE model, we implement a stepwise batch size adjustment—transitioning through 2M, 4M, 5M, and 6M at 125B-token intervals. In this configuration, the learning rate is scaled synchronously according to the square-root rule ($\eta \propto \sqrt{B}$). This approach is then compared against our primary experimental setup, which employs dynamic batch size adjustment while maintaining a constant learning rate.The training result is shown in Figure \ref{fig:loss_stable_scale}.

\begin{figure}
    \centering
    \includegraphics[width=1.0\linewidth]{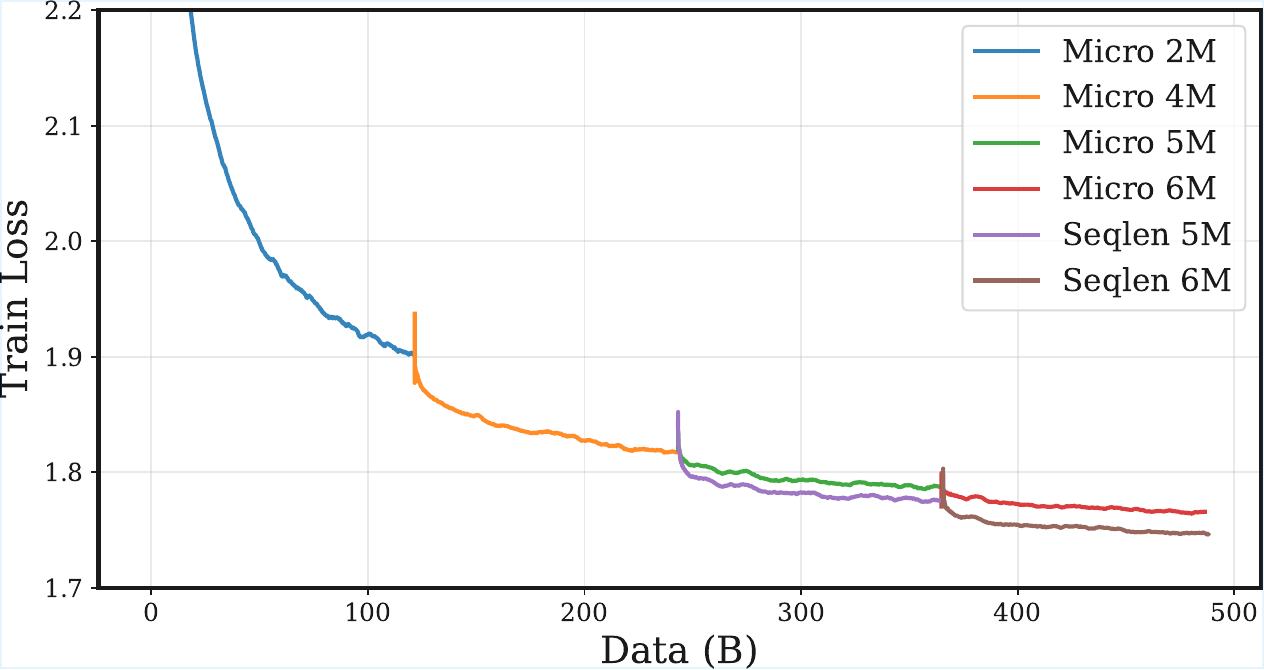}
    \caption{Comparison of training loss trajectories for dynamic batch size scaling through micro-batch adjustment and sequence length scaling.}
    \label{fig:loss_seqlen}
\end{figure}

\noindent \textbf{Switching the Sequence Length}
We conducted a comparative ablation study using the Qwen3 MoE model to investigate the impact of sequence length scaling. In the baseline configuration, the sequence length (\textit{seqlen}) was fixed at 4K, with 125B tokens processed for each batch size stage: 2M, 4M, 5M, and 6M. For the experimental group, we transitioned the \textit{seqlen} to 5K upon reaching the 250B-token mark and further increased it to 6K at 375B tokens. These adjustments resulted in batch sizes of 5M and 6M, respectively, effectively maintaining parity with the baseline's batch size trajectory while varying the underlying sample composition. The training result is shown in Figure \ref{fig:loss_seqlen}.

\begin{figure}
    \centering
    \includegraphics[width=1.0\linewidth]{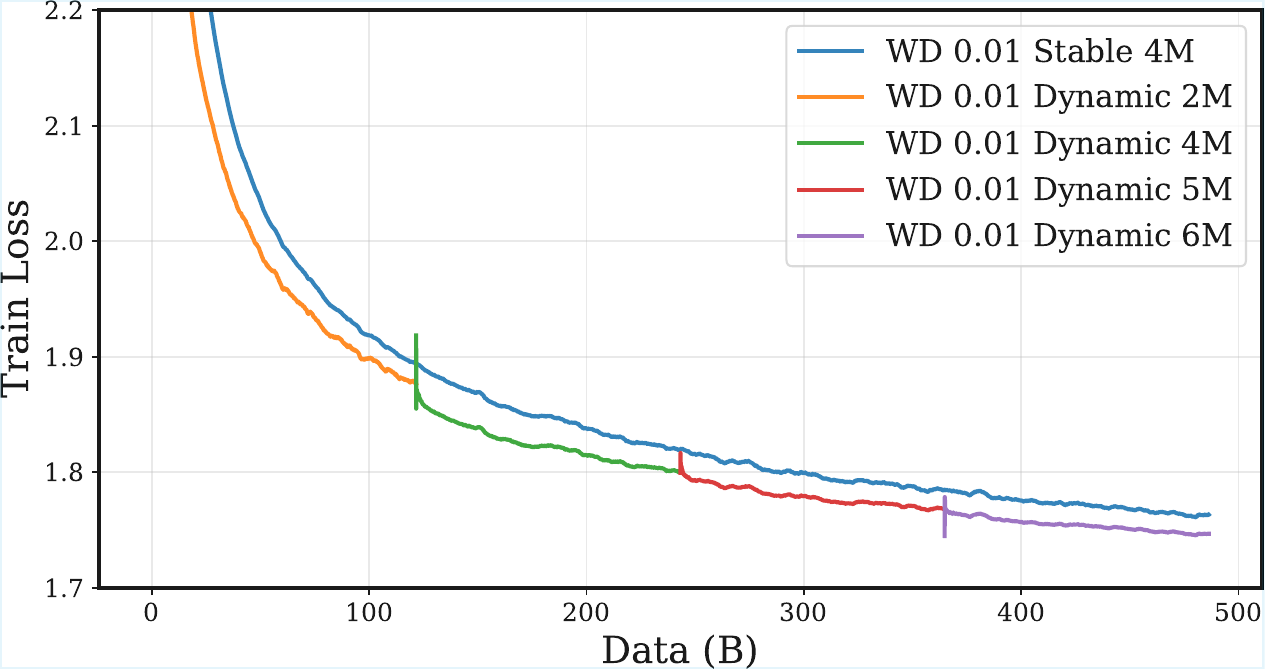}
    \caption{Comparison of training loss trajectories for fixed and dynamic batch size scheduling under different weight decay settings.}
    \label{fig:loss_wd}
\end{figure}

\noindent \textbf{Weight Decay}
We performed an ablation study based on the Qwen3 MoE setup, fixing the weight decay at 0.01 to compare the constant batch size regime against the dynamic batch size adjustment strategy. The training result is shown in Figure \ref{fig:loss_wd}.

\begin{figure}
    \centering
    \includegraphics[width=1.0\linewidth]{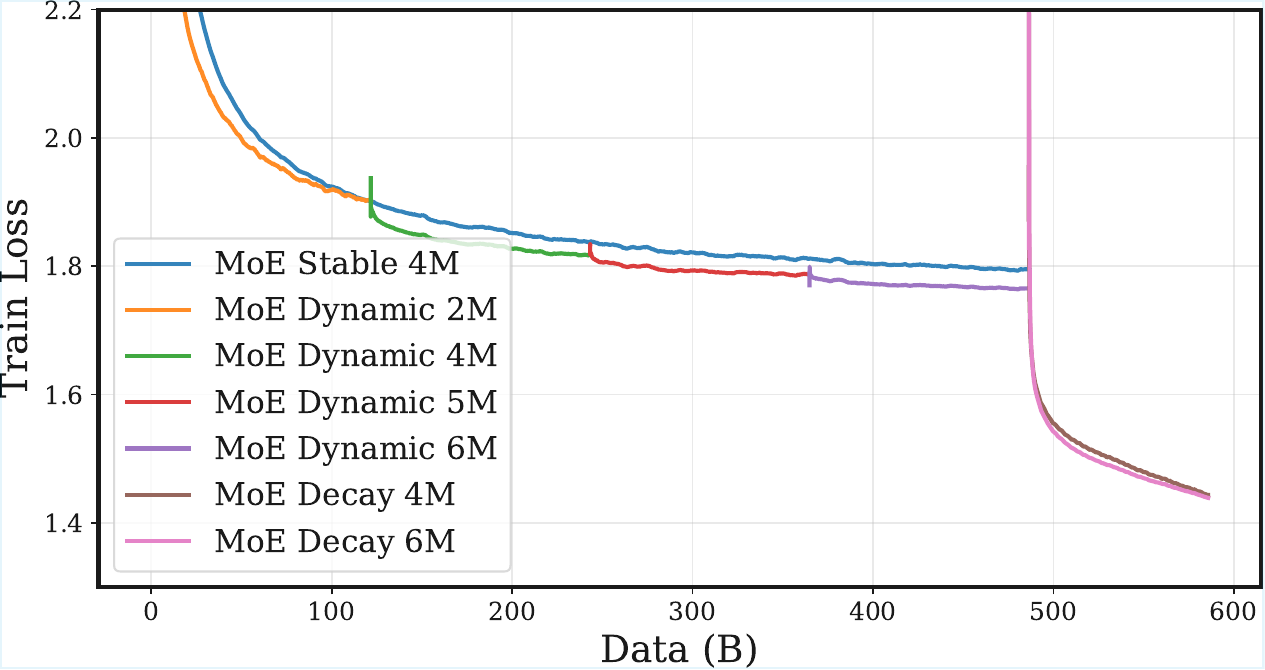}
    \caption{Training loss curves comparing the fixed and dynamic batch size strategies during continued training.}
    \label{fig:loss_stable_decay}
\end{figure}

\noindent \textbf{Continued Training}
Building upon the MoE model from the main experiments, we introduce an additional training phase characterized by learning rate decay. 
In this phase, the learning rate is linearly annealed to 10\% of the value used in the stable stage. 
The training is conducted over 100 billion (100B) tokens, utilizing the specialized data curated for the decay stage of InternLM2. Regarding the batch size settings, we fix the batch size at 4M for the baseline, whereas it is set to 6M for the dynamic batch size strategy. 
The experimental result is illustrated in Figure~\ref{fig:loss_stable_decay}.




\end{document}